\documentclass[a4paper,fleqn]{cas-sc} 

\usepackage[numbers,sort&compress]{natbib}
\usepackage{setspace}

\usepackage{caption}
\usepackage{booktabs}
\usepackage{graphicx}
\usepackage{subcaption}  
\usepackage{multirow}
\usepackage{float}
\usepackage{tabularx}
\usepackage{placeins}
\usepackage{booktabs, makecell, hhline}%
\usepackage{algorithm}
\usepackage{algpseudocode}
\usepackage{xcolor}
\definecolor{xl}{RGB}{0,102,204}

\def\tsc#1{\csdef{#1}{\textsc{\lowercase{#1}}\xspace}}
\tsc{WGM}
\tsc{QE}
\tsc{EP}
\tsc{PMS}
\tsc{BEC}
\tsc{DE}
\begin{document}
\let\WriteBookmarks\relax
\def\floatpagepagefraction{1}
\def\textpagefraction{.001}

\shortauthors{M. Tan et~al.}

\title [mode = title]{Self-Supervised Federated Learning under Data Heterogeneity for Label-Scarce Diatom Classification}                      
\author[1]{Mingkun Tan}[orcid=0000-0002-5997-9087]
\ead{mtan@cebitec.uni-bielefeld.de}
\cormark[1]
\author[2]{Xilu Wang}
\author[3]{Michael Kloster}
\author[1]{Tim W. Nattkemper}[orcid=0000-0002-7986-1158]
\cormark[1]
\ead{tim.nattkemper@uni-bielefeld.de}

\affiliation[1]{organization={Biodata Mining Group, Faculty of Technology, University of Bielefeld},
                country={Germany}}

\affiliation[2]{organization={Computer Science Research Centre, University of Surrey},
                country={United Kingdom}}

\affiliation[3]{organization={Phycology Group, Faculty of Biology, 
University of Duisburg-Essen},
                country={Germany}}

\cortext[cor1]{Corresponding authors}

\begin{abstract} 
Label-scarce visual classification under decentralized and heterogeneous data is a fundamental challenge in pattern recognition, especially when sites exhibit partially overlapping class sets. While self-supervised federated learning (SSFL) offers a promising solution, existing studies commonly assume the same data heterogeneity pattern throughout pre-training and fine-tuning. Moreover, current partitioning schemes often fail to generate pure partially class-disjoint data settings, limiting controllable simulation of real-world label-space heterogeneity. In this work, we introduce SSFL for diatom classification as a representative real-world instance and systematically investigate stage-specific data heterogeneity. We study cross-site variation in unlabeled data volume during pre-training and label-space misalignment during downstream fine-tuning. To study the latter in a controllable setting, we propose \emph{PreDi}, a partitioning scheme that disentangles label-space heterogeneity into two orthogonal dimensions, namely class \textbf{Pre}valence and class-set size \textbf{Di}sparity, enabling separate analysis of their effects. Guided by the resulting insights, we further propose \emph{PreP-WFL} (\textbf{Pre}valence-based \textbf{P}ersonalized \textbf{W}eighted \textbf{F}ederated \textbf{L}earning) to adaptively strengthen rare-class representations in low-prevalence scenarios. Extensive experiments show that SSFL consistently outperforms local-only training under both homogeneous and heterogeneous settings. The pronounced heterogeneity in unlabeled data volume is associated with improved representation pre-training, whereas under label-space heterogeneity, prevalence dominates performance and disparity has a smaller effect. PreP-WFL effectively mitigates this degradation, with gains increasing as prevalence decreases. These findings provide a mechanistic basis for characterizing label-space heterogeneity in decentralized recognition systems.
\end{abstract}
\maketitle
\section{Introduction}
Reliable visual classification remains difficult when data are distributed across isolated sites with raw-data sharing restrictions and significant data heterogeneity. Unlike large-scale natural image benchmarks such as ImageNet~\cite{deng2009imagenet}, expert-labeled domains, including medical imaging and biological taxonomy, impose restrictive annotation requirements that demand specialized domain knowledge. Consequently, the high acquisition cost of reliable labels establishes label scarcity as a primary bottleneck. This difficulty is further amplified in decentralized settings, where distributed clients exhibit substantial cross-client heterogeneity. Diatom classification provides a representative real-world instance of this problem. Automated taxonomic identification in this domain is constrained by limited expert availability, while imagery collected by different institutions and at different sites (and/or times) naturally exhibit heterogeneous structure due to geographical and temporal variations (like seasonal patterns) in species distributions, differences in sampling protocols, and unequal resource availability~\cite{su2024research}. As a result, client datasets often vary substantially in data volume and may contain only partially overlapping label spaces, reflecting a level of heterogeneity that exceeds the conventional federated learning assumption that clients share the same label space and differ only in label imbalance.

As high-resolution bioindicators for environmental assessment and global monitoring, diatom communities exhibit remarkable sensitivity to ecological shifts, making their composition a critical metric for ecosystem health \cite{ lobo2016diatoms, blanco2024diatom, henson2021future}. Recent advances in deep learning have significantly improved automated diatom classification~\cite{carcagni2021investigation,kloster2020deep,pu2023microscopic, yong2022comparison}. However, most existing approaches are supervised and therefore depend on large, expertly annotated datasets. Various strategies have been explored to alleviate this label scarcity, including annotation-assistance and active learning frameworks~\cite{kloster2020deep,langenkamper2017biigle, Moeller2021almi,haug2021combined}, data augmentation~\cite{liu2018teaching,vallez2022diffeomorphic,wang2017cgan}, few-shot learning~\cite{bueno2025phytoplankton, guo2021classification}, semi-supervised approaches~\cite{kloster2020deep,schmarje2021fuzzy}, and unsupervised clustering ~\cite{ibrahim2020image,pastore2023efficient}, yet each 
introduces specific trade-offs in annotation cost, scalability, robustness~\cite{tan2022impact}, or domain transferability. Among these directions, self-supervised learning (SSL) has emerged as a particularly promising paradigm for label-efficient representation learning. By learning representations from large volumes of unlabeled data and transferring them to downstream fine-tuning with limited annotations, SSL substantially reduces reliance on expert labels~\cite{chen2020simple,he2022masked}. 

Recent studies in diatom and plankton classification provide strong evidence for the effectiveness of SSL under annotation-scarce conditions. Tan et al.~\cite{tan2025scaling} reported that SSL pre-trained models yield larger relative gains as labeled data diminish. Notably, fine-tuning with only 50 samples per class achieves accuracy comparable to fully supervised training, thereby reducing annotation requirements by approximately 96\%. Similarly, Kareinen et al.~\cite{kareinen2025self} demonstrated that SSL pre-training on a large, diverse plankton corpus enables generalizable representations across different instruments and species distributions, revealing an inverse relationship between labeled data volume and the performance benefit of SSL. However, these SSL pipelines assume centralized access to all training data, which is a critical limitation given that valuable diatom and plankton imagery is inherently distributed across institutions, field stations, monitoring networks, and geographical regions~\cite{lombard2019globally}.

Federated learning (FL) addresses this decentralization challenge by enabling collaborative training without sharing raw data~\cite{li2020federated, mcmahan2017communication}. While the feasibility of FL has been demonstrated for specialized tasks such as phytoplankton classification~\cite{zhang2022plankton}, conventional supervised FL pipelines remain dependent on annotated datasets and therefore fail to fully exploit the large volume of unlabeled data distributed across collaborative networks. Self-supervised federated learning (SSFL), which naturally combines SSL with FL, offers a promising solution by enabling representation learning from decentralized unlabeled data.  However, its effectiveness in the diatom domain has not been systematically studied. In addition, existing SSFL studies typically assume the same heterogeneity pattern across both self-supervised pre-training and downstream fine-tuning stages~\cite{yan2023label}. Such a design overlooks the fact that heterogeneity may arise in different forms at different stages and may affect performance in fundamentally different ways. During pre-training, the primary challenge often lies in unequal unlabeled data volume across institutions, whereas during fine-tuning, the critical issue may shift to label-space heterogeneity, with clients holding only partially overlapping taxonomic label sets. Assuming these stage-specific data heterogeneities are the same pattern obscures their individual effects and hinders principled SSFL system design.

Moreover, conventional FL benchmarks typically characterize heterogeneity within a shared label space. In many real-world decentralized recognition problems, however, institutions observe only partially overlapping label spaces in addition to differing sample sizes. This form of heterogeneity is qualitatively more challenging, because the federation must integrate locally learned predictors over structurally misaligned class spaces rather than merely rebalancing class proportions. Recent FL studies have explored settings involving label-set mismatch and partially class-disjoint data (PCDD), and have proposed corresponding algorithmic solutions across diverse recognition tasks~\cite{xu2023federated,deng2023scale,fan2023FedMR,fan2023FedGELA,guo2025exploring}.  Nevertheless, the partitioning schemes used in these works, commonly based on Dirichlet sampling~\cite{guo2025exploring}, pathological partitioning~\cite{deng2023scale, fan2023FedGELA}, or task-specific masking~\cite{xu2023federated}, do not provide controllable and genuinely pure PCDD constructions~\cite{fan2023FedMR}. Dirichlet-based schemes yield missing classes only indirectly through extreme label skew, while fixed-class-per-client designs conflate how widely a class is shared across clients with how many classes each client contains. As a result, the underlying factors of label-space heterogeneity remain entangled, making it difficult to provide controllable and realistic simulations of real-world label-space heterogeneity. 

To address these challenges, we introduce an SSFL framework for diatom classification under stage-specific data heterogeneity: cross-client variation in unlabeled data volume during pre-training and label-space misalignment during downstream fine-tuning. The proposed framework leverages distributed unlabeled data for representation learning without sharing raw images and requires only 50 labeled samples per class for fine-tuning. Beyond establishing the feasibility of SSFL in this application, this work seeks to provide a mechanistic understanding of how stage-specific data heterogeneity affects performance. To this end, we analyze the two forms of heterogeneity separately. To fill the methodological gap in controllable partitioning, we propose \emph{PreDi}, a partitioning scheme that disentangles label-space heterogeneity into two orthogonal dimensions: Prevalence and Disparity. Guided by the resulting insights, we further develop \emph{PreP-WFL}, a prevalence-based personalized strategy that adaptively strengthens rare-class representations during local fine-tuning while preserving data locality. The main contributions of this work are as follows:
\begin{itemize}
       \item \textbf{Systematic evaluation of SSFL under stage-specific heterogeneous multi-client data.} We evaluate SSFL for diatom classification under independent and identically distributed (IID) settings and two non-IID settings: (i) heterogeneity in unlabeled data volume, where institutions possess different amounts of unlabeled data during self-supervised pre-training, and (ii) heterogeneity in labeled class space, where institutions exhibit distinct taxonomic class distributions during downstream fine-tuning. The results show that SSFL consistently outperforms local-only training under both homogeneous and heterogeneous settings.

    \item \textbf{PreDi: a two-dimensional partitioning scheme for label-space heterogeneity.} We propose \textit{PreDi}, a partitioning scheme that characterizes label-space heterogeneity along two complementary dimensions: class \textbf{Pre}valence, defined as the number of clients in which a given class appears, and class-set size \textbf{Di}sparity, defined as the variation in the number of classes across clients. This disentangled design enables controlled analysis of their individual effects and reveals prevalence as the primary factor affecting SSFL performance under label-space heterogeneity.
    
    \item \textbf{PreP-WFL: prevalence-based personalized weighting for low-prevalence settings.} We propose \textit{PreP-WFL}, a \textbf{Pre}valence-based \textbf{P}ersonalized \textbf{W}eighted \textbf{F}ederated \textbf{L}earning scheme that adaptively adjusts class weights during local fine-tuning according to global cross-client prevalence. Extensive experiments show that PreP-WFL mitigates performance degradation in challenging low-prevalence regimes, with gains increasing as prevalence decreases.
\end{itemize}

\begin{figure}
	\centering
	\includegraphics[width=\textwidth]{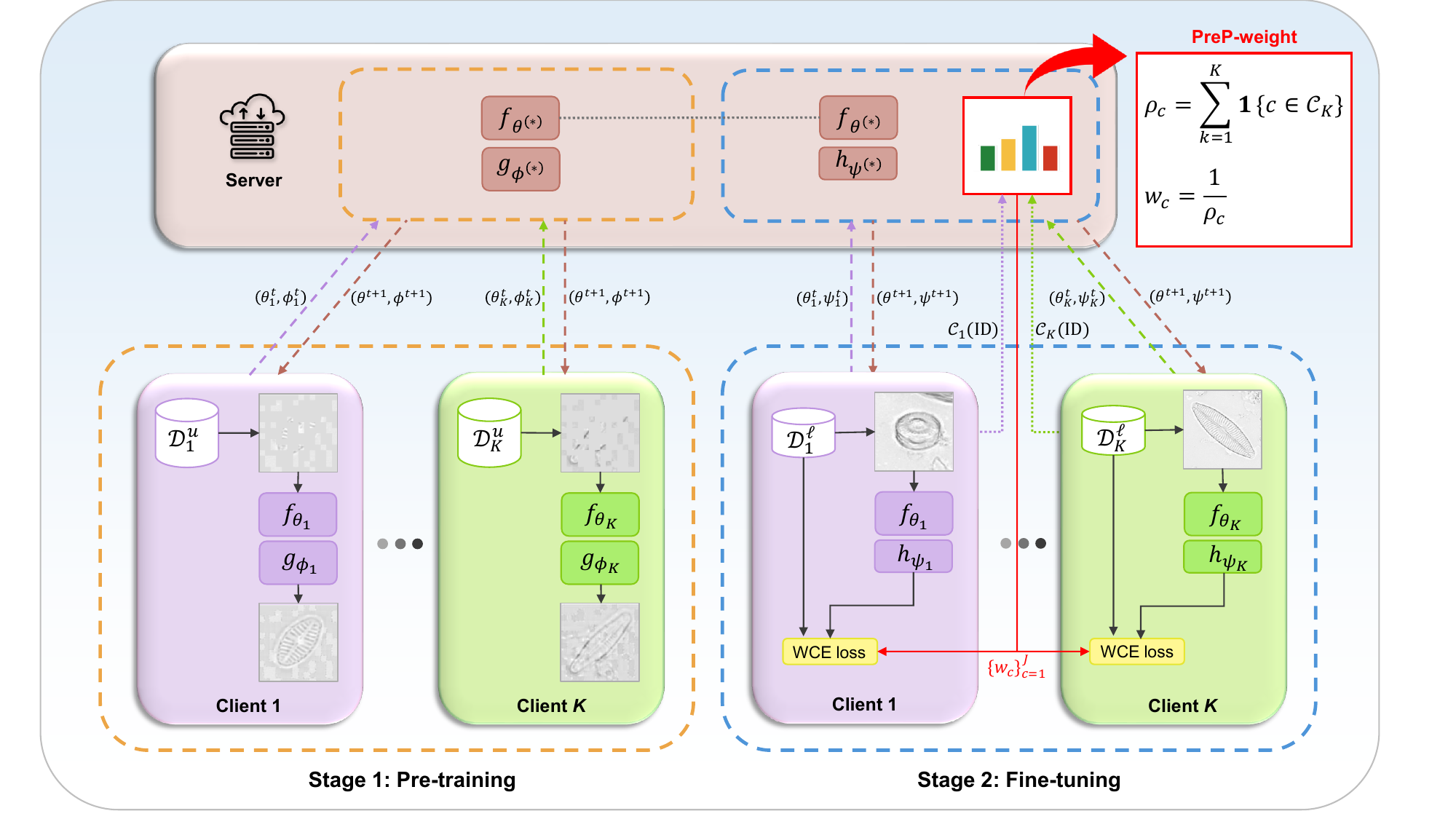}
	\caption{Overview of the self-supervised federated learning framework with PreP-WFL. 
    In the pre-training stage, masked image reconstruction is adopted as the self-supervised task: at each communication round $t$, every client $k\in\{1,\dots,K\}$ updates its local encoder $f_{\theta_k}$ and decoder $g_{\phi_k}$ on the unlabeled dataset $\mathcal{D}^u_k$, and uploads the parameters $(\theta^t_k,\phi^t_k)$ to the server. The server performs FedAvg to obtain a global encoder $(f_{\theta^{(*)}},g_{\phi^{(*)}})$ and broadcasts it to all clients. In the fine-tuning stage, the final global encoder $f_{\theta^{(*)}}$ initializes the local encoders, and a classifier head $h_{\psi_k}$ is attached on each client and trained on its labeled dataset $\mathcal{D}^\ell_k$. The PreP\mbox{-}weight module (red box) performs a one-shot aggregation of the clients’ label sets ${\mathcal{C}_k}$ \emph{(label IDs only, no images)} to compute the global class prevalence $\rho_c$ and the corresponding weights $w_c$. These weights are then broadcast to all clients and incorporated into a weighted cross-entropy loss during local fine-tuning, thereby emphasizing rare, site-specific taxa in the federated model.}
	\label{FIG:1}
\end{figure}
\section{Related work}
\subsection{Self-supervised learning}
Self-supervised learning (SSL) has become a major approach for label-efficient visual representation learning by exploiting supervisory signals derived from unlabeled data itself~\cite{misra2020self}. Existing SSL methods can be broadly divided into joint-embedding and self-prediction paradigms. Joint-embedding methods, including contrastive approaches such as SimCLR~\cite{chen2020simple} and MoCo~\cite{he2020momentum}, learn representations by pulling closer positive pairs and pushing apart negative pairs, typically requiring large batch sizes or memory banks. Negative-free variants such as BYOL~\cite{grill2020bootstrap} and SimSiam~\cite{chen2021exploring} instead predict one augmented view from another of the same image without explicit negative samples. Self-prediction methods include earlier tasks such as innate relationship prediction~\cite{xu2019self} and more recent generative schemes that reconstruct masked or corrupted regions. Representative examples are BEiT~\cite{bao2021beit} and masked autoencoder (MAE)~\cite{he2022masked}, commonly referred to as masked image modeling methods, which have shown strong transferability across visual recognition tasks.

Recent work on plankton and diatom imagery has shown that in-domain MAE pre-training consistently outperforms conventional ImageNet initialization, especially in low-label regimes~\cite{kareinen2025self,ciranni2025domain,tan2025scaling}. In marine plankton recognition, Kareinen et al.~\cite{kareinen2025self} showed that in-domain MAE pre-training on large-scale unlabeled plankton imagery yields consistently higher classification accuracy than conventional ImageNet pre-training, particularly in the low-label regime, underscoring the advantage of domain-specific SSL under distribution shift. Ciranni et al.~\cite{ciranni2025domain} further analyzed in-domain SSL under both computational and labeling constraints. They show that MAE-based pre-training can yield better representations for downstream classification than ImageNet pre-training even when the in-domain plankton dataset is smaller, and that competitive performance can be achieved with a lightweight classifier trained on top of frozen features. In the diatom domain, Tan et al.~\cite{tan2025scaling}, building on the large-scale UDE diatoms in the Wild 2024 dataset~\cite{venkataramanan2024ude}, systematically evaluated MAE and found that it substantially improves data utilization efficiency. In particular, its relative gains increase as the available labeled dataset size decreases, and the pre-trained backbone can achieve accuracy comparable to fully supervised baselines with as few as \(50\) labeled samples per class, highlighting the annotation efficiency of MAE in specialized diatom image analysis.

These studies establish SSL as a strong foundation for label-scarce diatom classification. However, existing studies assume centralized access to all unlabeled data and therefore do not address the decentralized multi-institution setting considered in this work.

\subsection{Federated learning and data heterogeneity}
Federated learning (FL) enables collaborative model training across distributed clients without sharing raw data~\cite{kairouz2021advances,li2020federated}. In the standard pipeline, each client performs local optimization on its own dataset and transmits model updates to a central server, which aggregates them into a global model, most commonly using Federated Averaging (FedAvg)~\cite{mcmahan2017communication}. A central challenge in FL is data heterogeneity. In practice, client datasets are rarely IID and often differ in sample size, class distribution, feature characteristics, and acquisition conditions~\cite{li2020federated,su2024research,gao2022survey}. Such non-IID structure can impair convergence, bias aggregation toward dominant clients, and reduce generalization of the global model~\cite{li2020federated,lu2024federated}. A broad range of methods has been proposed to address this~\cite{li2020federated,fallah2020personalized,
li2021ditto}, but these studies predominantly assume a homogeneous global label space in which every client can in principle observe any class.

However, this assumption breaks down whenever clients operate in specialized or geographically constrained environments. In multi-institution ecological monitoring, for instance, geographically dispersed sites observe only site-specific subsets of species, naturally producing partially overlapping or even disjoint label sets across the federation. This form of heterogeneity poses a qualitatively harder problem than class-proportion imbalance within a shared label space.  Rather than rebalancing sample frequencies, federated aggregation must integrate classifiers trained on structurally incompatible output spaces, causing locally missing classes to receive degraded or entirely absent representations. Zhang et al.~\cite{zhang2022plankton} demonstrated the feasibility of FL for phytoplankton classification and observed performance degradation under non-IID data, but their setting remained within the conventional shared-label-space assumption and did not address this more severe form of structural mismatch.  

Recent works have proposed algorithmic solutions for label-space heterogeneity across diverse recognition settings~\cite{xu2023federated,deng2023scale,fan2023FedMR,fan2023FedGELA, guo2025exploring}. However, a closer examination of how these works construct their experimental partitions, reveals a persistent methodological limitation. Fed-MENU \cite{xu2023federated} restricts each site to annotating only a predefined subset of target structures in a segmentation task, producing existential label gaps through a task-specific masking protocol. FedLSM~\cite{deng2023scale} randomly distributes training samples across clients and then designates a fixed number of classes per client as locally known, treating the remainder as locally unknown. FedVLS \cite{guo2025exploring} applies Dirichlet sampling $\text{Dir}(\alpha)$ with a small concentration parameter $\alpha$, which probabilistically produces vacant classes at individual clients as a byproduct of strong label skew. FedGELA \cite{fan2023FedGELA} and rely on pathological class assignment, randomly allocating a fixed subset of classes to each client to simulate disjointness.

However, Dirichlet-based partitioning, as explicitly noted by FedMR~\cite{fan2023FedMR}, fails to produce a pure PCDD setting; FedMR therefore proposes a dedicated $P_{\rho}C_{\varsigma}$ protocol, which divides the dataset across $\rho$ clients and each client holds $\varsigma$ classes. This design indeed produces a purer PCDD construction than conventional Dirichlet partitioning. Nevertheless, a critical issue remains unresolved. Although the protocol controls how many classes each client contains, it does not explicitly disentangle this factor from how widely each class is shared across clients. Consequently, these two distinct dimensions of label-space heterogeneity are typically varied together, making it difficult to determine whether performance degradation, or any algorithmic gain, is primarily driven by one factor, the other, or their interaction.

\subsection{Self-supervised federated learning}
Self-supervised federated learning (SSFL) combines the annotation efficiency of SSL with the data-local and collaborative properties of FL. The standard pipeline typically proceeds in two sequential phases: self-supervised representation pre-training on distributed unlabeled data, followed by supervised fine-tuning on limited labeled samples. This paradigm has attracted growing interest in settings where annotation is costly and raw-data sharing is restricted. Preliminary SSFL frameworks have been applied within the medical imaging domain, establishing the paradigm's efficacy for annotation-scarce recognition under conventional non-IID constraints. Yan et al.~\cite{yan2023label} proposed a Transformer-based federated masked image modeling framework and demonstrated that Fed-MAE achieves superior robustness over Fed-BEiT under heterogeneous data distributions, particularly in severely non-IID scenarios. Khowaja et al.~\cite{Ali_Khowaja_2025} introduced SelfFed, which augments the federated pipeline with a contrastive network and a heterogeneity-aware aggregation strategy to further alleviate label scarcity during fine-tuning. 

However, a common implicit assumption of existing SSFL studies is that the same data heterogeneity pattern persists throughout pre-training and downstream fine-tuning. In most benchmarks, a dataset is partitioned once, and the same non-IID pattern is used in both stages. This design overlooks the fact that different forms of heterogeneity may arise at different stages of the SSFL pipeline and may affect performance in different ways. During pre-training, the relevant challenge is unlabeled data volume, since clients may differ substantially in how much unlabeled imagery they can contribute to self-supervised representation learning. Centralized SSL studies have established that pre-training quality is sensitive to data scale~\cite{tan2025scaling,
kareinen2025self}, yet how unequal client contributions affect SSFL pre-training remains unexplored. During fine-tuning, the main challenge in our setting shifts to label-space heterogeneity, as clients may possess only partially overlapping or even disjoint taxonomic label sets. Conflating these stage-specific data heterogeneities may obscure their individual effects and impede principled system design.

To the best of our knowledge, no existing SSFL study decouples unlabeled volume heterogeneity during pre-training from label-space heterogeneity during fine-tuning, nor systematically analyzes their individual effects on downstream recognition performance. Our work addresses these gaps by assessing stage-specific data heterogeneity sources independently, and introduces PreDi to further disentangle label-space heterogeneity into its constituent factors. Subsequently, we propose PreP-WFL to mitigate the dominant source of performance degradation identified through this analysis.

\section{Methods}
In this section, we first describe the reconstruction of the dataset used in our experiments (\S\ref{sec:dataset}). We then present the SSFL paradigm adopted in this study (\S\ref{sec:ssflform}). Next, we introduce a controlled construction protocol for stage-specific data heterogeneity, including non-IID unlabeled data partitioning and non-IID labeled data partitioning based on PreDi (\S\ref{sec:heterogeneity}). Finally, we propose PreP-WFL, a prevalence-based personalized weighting strategy designed to improve robustness in low-prevalence regimes during local fine-tuning (\S\ref{sec:prepwfl}).
\subsection{Dataset reconstruction}
\label{sec:dataset}

\begin{figure}
	\centering
	\includegraphics[width=.7\textwidth]{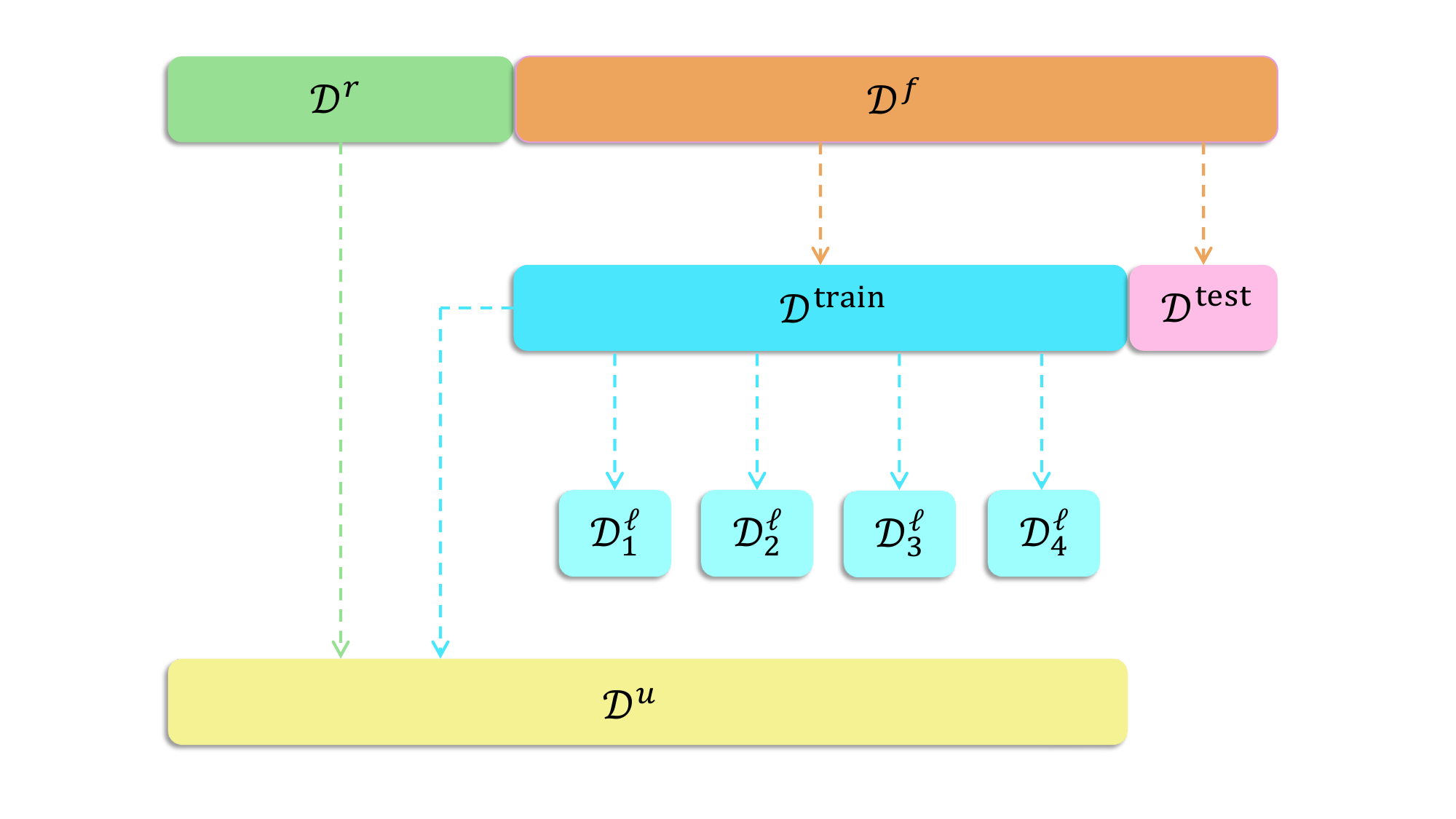}
	\caption{Overview of the dataset reconstruction pipeline. From the original dataset, we first derive the filtered labeled set $\mathcal{D}^{f}$, which is partitioned into training and test sets $\mathcal{D}^{\mathrm{train}}$ and $\mathcal{D}^{\mathrm{test}}$ (\(80/20\)). The remaining taxa form $\mathcal{D}^r$. From $\mathcal{D}^{\mathrm{train}}$, we sample balanced (50 images per class) labeled subsets $\{\mathcal{D}^{\ell}_k\}_{k=1}^{4}$ to serve as the labeled dataset on each client for downstream fine-tuning. The unlabeled pool $\mathcal{D}^{u}$ is constructed by combining $\mathcal{D}^{\mathrm{train}}$ with $\mathcal{D}^r$ and discarding all labels.}
	\label{fig:dataset}
\end{figure}

We use the UDE diatoms in the Wild 2024 dataset~\cite{venkataramanan2024ude}, the largest real-world diatom image dataset to date, comprising \(|\mathcal{D}^{\text{all}}| = 83{,}570\) images from \(611\) taxa. An overview of the resulting construction
is illustrated in Fig.~\ref{fig:dataset}. To ensure sufficient sample size for statistically reliable evaluation, we retain only taxa with at least \(350\) images. This yields
a filtered labeled set \(\mathcal{D}^{f} \subset \mathcal{D}^{\text{all}}\) with
\(|\mathcal{D}^{f}| = 55{,}542\) images from \(C = 36\) classes:
\(
\mathcal{D}^{f} = \{(x_i, y_i)\}_{i=1}^{55{,}542}, \quad 
y_i \in \{1,\dots,C\}.
\) We then randomly split \(\mathcal{D}^{f}\) into training and test sets,
\(\mathcal{D}^{\text{train}}\) and \(\mathcal{D}^{\text{test}}\), using an \(80/20\) ratio.
Let \(\mathcal{D}^{r} = \mathcal{D}^{\text{all}} \setminus \mathcal{D}^{f}\) denote
all images from taxa with fewer than \(350\) samples. The unlabeled pool
\(\mathcal{D}^u\) is then constructed by merging \(\mathcal{D}^{\text{train}}\) with \(\mathcal{D}^{r}\) and
discarding all labels:
\(
\mathcal{D}^u = \{x \mid (x,y) \in \mathcal{D}^{\text{train}} \cup \mathcal{D}^{r}\}.
\)

To simulate a real-world FL setting, we consider \(K = 4\) clients, each holding unlabeled and labeled data. Following Tan et al.~\cite{tan2025scaling}, we create a label-scarce, balanced subset for downstream fine-tuning by sampling \(n^{\ell} = 50\) distinct labeled images per class from \(\mathcal{D}^{\text{train}}\) to form four labeled subsets
\(\{\mathcal{D}^{\ell}_k\}_{k=1}^{4}\), each covering all \(C = 36\) classes with
\(|\mathcal{D}^{\ell}_k\}| = C \cdot n^{\ell} = 36 \times 50\) labeled images:
\(
\mathcal{D}^{\ell}_k = \{(x_i^{(k)}, y_i^{(k)})\}_{i=1}^{36 \times 50},
\quad
y_i^{(k)} \in \{1,\dots,36\}.
\)
These labeled subsets and the unlabeled pool are subsequently partitioned to simulate various forms of data heterogeneity across clients; for each client \(k\), the local dataset is
\(\mathcal{D}_k = \mathcal{D}^{u}_k \cup \mathcal{D}^{\ell}_k\). We next formalize the proposed SSFL setting and then describe how the two forms of client heterogeneity are simulated.

\subsection{SSFL Paradigm}
\label{sec:ssflform}

We study label-scarce image classification in a decentralized multi-client setting, where raw data remain local to participating institutions. Following the Fed-MAE paradigm, we adopt a two-stage SSFL pipeline. In the first stage, clients collaboratively perform federated self-supervised pre-training on distributed unlabeled data to learn a shared encoder. In the second stage, the resulting global encoder is transferred to downstream supervised adaptation using limited labeled samples available at each client.

Formally, let \(K\) denote the number of clients, and let client \(k \in \{1,\ldots,K\}\) hold a local unlabeled dataset \(\mathcal{D}_k^u = \{x_i^{(k)}\}_{i=1}^{n_k}\). During pre-training, a masking operator \(\mathcal{M}\) randomly drops a large fraction of image patches, yielding the visible subset
\[
x_{\mathrm{vis},i}^{(k)}=\mathcal{M}\big(x_i^{(k)}\big).
\]
A local encoder \(f_{\theta_k}\) and decoder \(g_{\phi_k}\) are then trained to reconstruct the masked regions. The local MAE objective is given by the reconstruction error over masked pixels:
\[
\mathcal{L}^{\text{MAE}}_k(\theta_k,\phi_k)
=
\mathbb{E}_{x^{(k)} \sim \mathcal{D}_k^u}
\left[
\left\|
g_{\phi_k}\big(f_{\theta_k}(x_{\mathrm{vis},i}^{(k)})\big)_{\Omega_{\mathrm{mask}}}
-
x^{(k)}_{i,\Omega_{\mathrm{mask}}}
\right\|_2^2
\right].
\]
After local optimization, clients upload model parameters to the server, which aggregates them using FedAvg:
\[
\theta^{(t+1)}=\sum_{k=1}^{K}\frac{n_k}{\sum_{j=1}^{K}n_j}\theta_k^{(t)},
\qquad
\phi^{(t+1)}=\sum_{k=1}^{K}\frac{n_k}{\sum_{j=1}^{K}n_j}\phi_k^{(t)}.
\]
The final global encoder \(f_{\theta^{(*)}}\) is then used to initialize downstream fine-tuning with local labeled datasets.

In the next subsection, we first simulate unlabeled-data-volume heterogeneity for pre-training, and then introduce PreDi to construct label-space heterogeneity data for the fine-tuning stage.

\subsection{Stage-specific data heterogeneity construction}
\label{sec:heterogeneity}
A key distinction between the setting considered here and conventional FL benchmarks lies in the structure of data heterogeneity. In many standard FL benchmarks, client-level non-IIDness is simulated primarily as label imbalance, i.e.,\ differences in class proportions across clients,
\(
p_k(y) \neq p_{k'}(y)\) for \(k \neq k'\), while the label space remains shared, $\mathcal{Y}_k = \mathcal{Y}_{k'} = \mathcal{Y}$. In such settings, heterogeneity mainly affects sample proportions within a common label space. 

By contrast, diatom classification involves rich taxonomic diversity with a large label space \(\mathcal{Y}\), and each institution \(k\) may observe only a subset \(\mathcal{Y}_k \subseteq \mathcal{Y}\) determined by its geographic region and habitat. Overlap between \(\mathcal{Y}_k\) and \(\mathcal{Y}_{k'}\) can be limited, with some taxa exclusive to individual sites, leading to pronounced label-space heterogeneity. This poses fundamentally different optimization challenges: federated aggregation must integrate information across partially disjoint label spaces rather than merely rebalancing samples within a shared taxonomy. In addition, the amount of unlabeled data available for self-supervised pre-training can vary substantially across institutions, ranging from large automated laboratories to small field stations. 

To study these effects in a controlled manner, we construct two forms of heterogeneity: cross-client variation in unlabeled data volume during SSFL pre-training and label-space heterogeneity in labeled data during downstream fine-tuning.

\subsubsection{Non-IID unlabeled data partitioning}
\begin{figure} 
\centering 
\includegraphics[width=.7\textwidth]{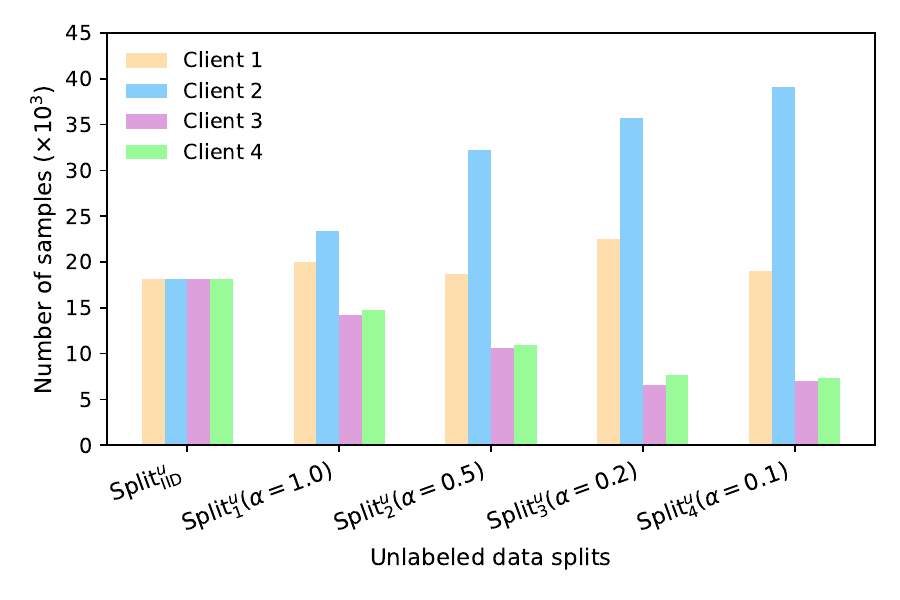} \caption{Heterogeneity in unlabeled data volume. 
Bars show the number of unlabeled samples per client for the IID
Split$^{u}_{\mathrm{IID}}$ and the four non-IID partitions 
Split$^{u}_{1}$–Split$^{u}_{4}$ (corresponding to $\alpha\!\in\!\{1.0,0.5,0.2,0.1\}$). 
See also Table~\ref{tab:pretrain_splits}.} 
\label{fig:pretrain_splits} 
\end{figure}

The IID split for unlabeled data is obtained by uniformly partitioning all unlabeled data across the four clients, yielding the split denoted by $\text{Split}^{u}_{\mathrm{IID}}$. To simulate quantity skew across clients, we adopt a standard Dirichlet-based sampling strategy. For $K$ clients and a total of $N$ unlabeled samples, we first draw a proportion vector \[
\mathbf{r} = (r_1, r_2, \dots, r_K)
\sim \mathrm{Dir}_K(\alpha),
\]
where $\alpha$ is the concentration parameter controlling the degree of skew in client data volumes: smaller values of $\alpha$ lead to more severe quantity skew. The number of unlabeled samples assigned to client $i$ is then given by
\[
n_i = N \cdot r_i, 
\quad \text{with } \sum_{i=1}^{K} n_i = N.
\]
As illustrated in Fig.~\ref{fig:pretrain_splits} (and Table~\ref{tab:pretrain_splits}), we construct four non-IID unlabeled data splits with
$\alpha \in \{1.0, 0.5, 0.2, 0.1\}$ for $K = 4$ clients. These partitions are denoted by
$\text{Split}^{u}_{1}$,
$\text{Split}^{u}_{2}$,
$\text{Split}^{u}_{3}$,
and $\text{Split}^{u}_{4}$.
\begin{table}[width=\linewidth,cols=5,pos=h]
\centering
\caption{Unlabeled data splits.
Each cell reports the number of unlabeled samples held by a given client.}
\label{tab:pretrain_splits}
\renewcommand{\arraystretch}{1.2}
\begin{tabular}{lcccc}
\toprule
\textbf{Splits}
& Client~1
& Client~2 
& Client~3 
& Client~4 \\
\midrule
\textbf{$\text{Split}^{u}_{\text{IID}}$}
  & 18,115 & 18,115 & 18,115 & 18,116 \\
\textbf{$\text{Split}^{u}_{1} (\alpha=1.0)$}
  & 19,988 & 23,405 & 14,258 & 14,810 \\
\textbf{$\text{Split}^{u}_{2} (\alpha=0.5)$}
  & 18,696 & 32,209 & 10,653 & 10,903 \\
\textbf{$\text{Split}^{u}_{3} (\alpha=0.2)$}
  & 22,558 & 35,750 & 6,546  & 7,607  \\
\textbf{$\text{Split}^{u}_{4} (\alpha=0.1)$}
  & 18,982 & 39,156 & 6,975  & 7,348  \\
\bottomrule
\end{tabular}
\end{table}

\subsubsection{Non-IID labeled data partitioning based on PreDi}
The IID split for labeled data corresponds to the case where each client holds all \(C=36\) classes and is denoted by \(\text{Split}^{\ell}_{\mathrm{IID}}\). Dirichlet-based non-IID partitioning controls skewness via a single
concentration parameter \(\alpha\). However, this scheme implicitly entangles two conceptually distinct aspects of label heterogeneity:
\begin{itemize}
    \item class prevalence: the number of clients in which a given class is present.
    \item class-set size disparity: the variation in the number of classes across clients.
\end{itemize}

\begin{figure}
    \centering
    \includegraphics[width=.8\textwidth]{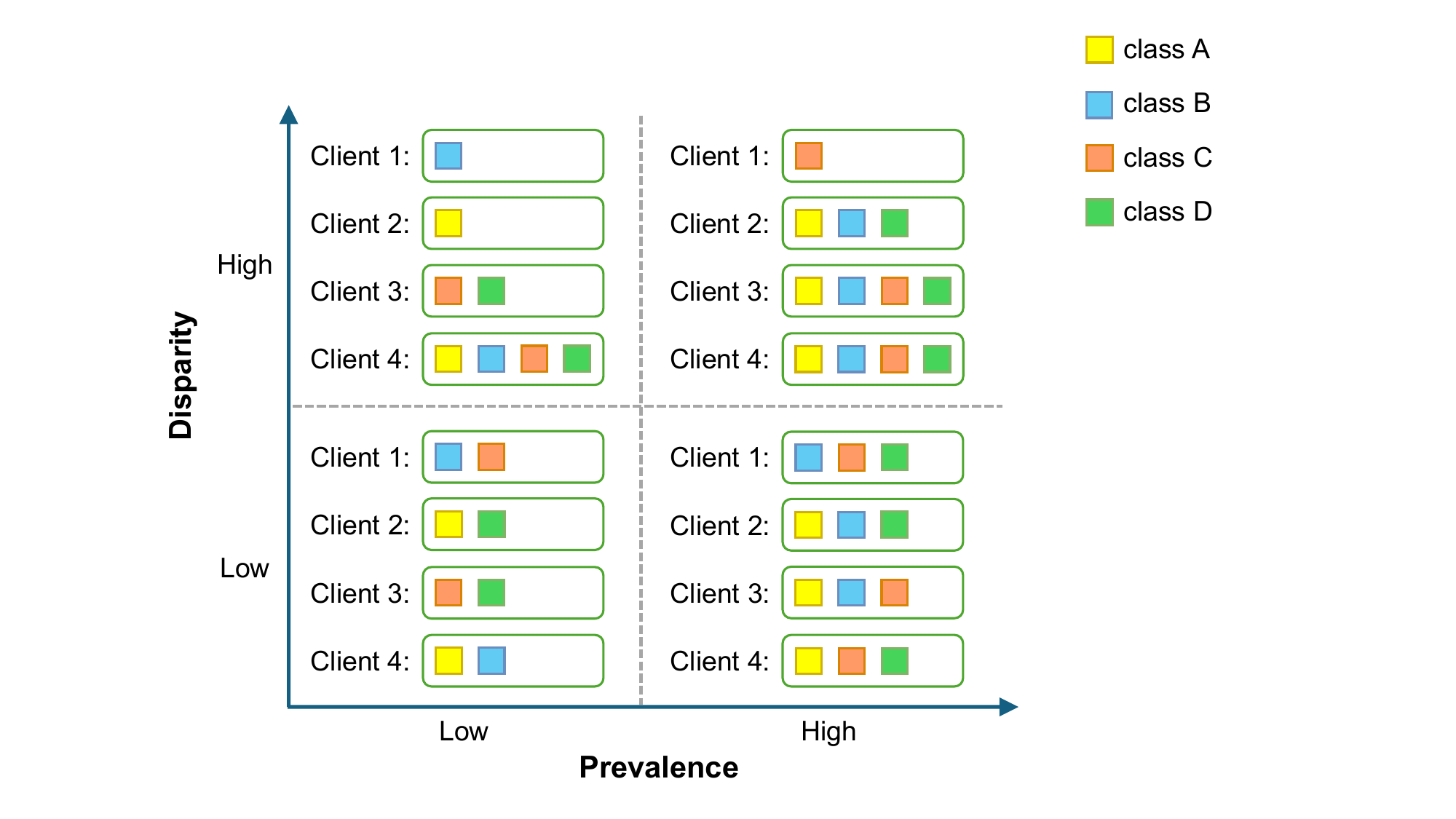}
    \caption{Illustration of class prevalence and class-set size disparity. 
    The x-axis indicates prevalence (low to high), the y-axis indicates disparity (low to high), and each colored circle denotes a class (A--D).}
    \label{fig:predi}
\end{figure}

As illustrated in Fig.~\ref{fig:predi}, the joint configuration of class prevalence and class-set size disparity gives rise to qualitatively different federation regimes. When prevalence is low, many classes are institution-specific, appearing in only one or a few clients; when prevalence is high, most classes are shared across institutions. Simultaneously, high disparity indicates that some institutions monitor a broad spectrum of classes while others observe only a small subset, whereas low disparity corresponds to clients exhibiting comparable numbers of classes. Different prevalence–disparity combinations correspond to distinct real-world monitoring scenarios, leading to different FL regimes. To explicitly disentangle these effects, we propose \emph{PreDi},
a partitioning scheme that decouples and independently controls class \textbf{Pre}valence and cross-client class-set size \textbf{Di}sparity, enabling a systematic analysis of how each component of label-space heterogeneity affects performance.

Let \(\mathcal{C} = \{1,\dots,C\}\) denote the set of classes and
\(\mathcal{K} = \{1,\dots,K\}\) the set of clients. We introduce a binary assignment matrix
\(\mathbf{M} \in \{0,1\}^{C \times K}\) with entries
\[
M_{c,k} =
\begin{cases}
1, & \text{if class } c \text{ is present at client } k,\\
0, & \text{otherwise.}
\end{cases}
\]
Based on \(\mathbf{M}\), we characterize label-space heterogeneity using two complementary statistics:
\begin{itemize}
    \item \textbf{average prevalence across classes}:
    \[
    \rho_c = \sum_{k=1}^{K} M_{c,k}, \quad
    \bar \rho = \frac{1}{C} \sum_{c=1}^{C} \rho_c,
    \]
    where \(\rho_c\) denotes the number of clients in which class \(c\) appears, and $\bar \rho$ is the average prevalence across all classes. A larger \(\bar \rho\) indicates that more classes are shared by more clients on average, whereas a smaller \(\bar \rho\) implies that more classes are confined to a small number of clients.
    
    \item \textbf{class-set size disparity across clients}:
    \[
    n_k = \sum_{c=1}^{C} M_{c,k}, \quad
    \mu = \frac{1}{K} \sum_{k=1}^{K} n_k, \quad
    \sigma = \sqrt{\frac{1}{K} \sum_{k=1}^{K} (n_k - \mu)^2},
    \]
    where \(n_k\) is the number of classes observed at client \(k\), \(\mu\) is the mean number of classes across clients, and \(\sigma\) is the corresponding population standard deviation. A larger \(\sigma\) indicates
    stronger cross-client disparity in class-set size, meaning that some clients contain many classes while others contain only a few.
\end{itemize}

We construct 20 non-IID labeled-data splits by specifying target values of average prevalence and disparity: \[
\bar{\rho}^* \in \{3.5, 3.0, 2.5, 2.0, 1.5\},
\sigma^* \in \{0.0, 1.0, 2.0, 3.0\}.
\]

\begin{figure}
    \centering
    \includegraphics[width=0.8\textwidth]{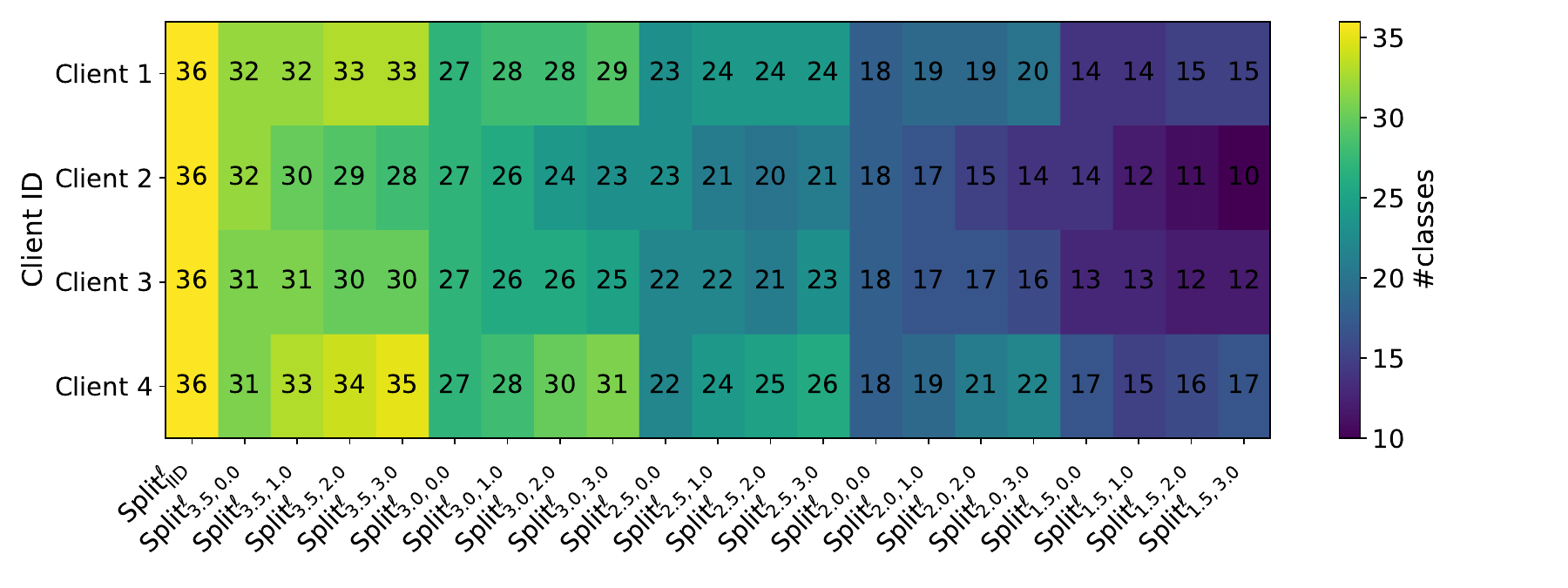}
    \caption{Heatmap of the number of classes per client under different labeled-data splits. The x-axis enumerates the IID split $\text{Split}^{\ell}_{\mathrm{IID}}$ and PreDi non-IID splits $\text{Split}^{\ell}_{\bar\rho^*,\sigma^*}$ obtained by varying $\bar\rho^* \in \{3.5, 3.0, 2.5, 2.0, 1.5\}$ and $\sigma^* \in \{0.0, 1.0, 2.0, 3.0\}$. The y-axis lists the clients, and each cell indicates the number of classes assigned to that client under the corresponding split.}
    \label{fig:labeled_noniid}
\end{figure}

Each target pair \((\bar{\rho}^*, \sigma^*)\) defines a desired federated configuration, denoted by
\(\text{Split}^{\ell}_{\bar{\rho}^*,\sigma^*}\). Because PreDi controls target statistics through a stochastic assignment process, the realized statistics of a generated split, namely \(\bar{\rho}\) and \(\sigma\), are generally close to, but not exactly equal to, their corresponding targets. For example, \(\text{Split}^{\ell}_{3.0,2.0}\) denotes a split generated with target values \(\bar{\rho}^*=3.0\) and \(\sigma^*=2.0\), whose realized statistics typically satisfy \(\bar{\rho} \approx 3.0\) and \(\sigma \approx 2.0\). The IID split and the 20 non-IID split configurations are summarized in Fig.~\ref{fig:labeled_noniid}. The PreDi method is detailed in Algorithm~\ref{alg:predi_algorithm}.

\begin{algorithm}[h]
\caption{PreDi: Prevalence--Disparity partitioning scheme}
\label{alg:predi_algorithm}
\begin{algorithmic}[1]
\Require Class set \(\mathcal{C}=\{1,\dots,C\}\), client set \(\mathcal{K}=\{1,\dots,K\}\), labeled dataset \(\mathcal{D}=\{(x_i,y_i)\}\), target statistics \((\bar{\rho}^*,\sigma^*)\), and per client-class sample size \(s=50\).
\Ensure Assignment matrix \(\mathbf{M}\in\{0,1\}^{C\times K}\) and client labeled datasets \(\{\mathcal{D}_k^\ell\}_{k=1}^K\).

\State \textbf{Prevalence assignment:} initialize \(\rho_c \gets 1\) for all \(c\), then randomly increase selected \(\rho_c\) values, subject to \(\rho_c \le K\), until \(\frac{1}{C}\sum_c \rho_c\) is close to \(\bar{\rho}^*\).

\State \textbf{Disparity assignment:} let \(A \gets \sum_c \rho_c\) and \(\mu \gets A/K\). Sample target class counts \(q_1,\dots,q_K\) with mean \(\mu\) and standard deviation close to \(\sigma^*\), then adjust them so that \(\sum_k q_k = A\).

\State \textbf{Class allocation:} initialize \(n_k \gets 0\) and \(M_{c,k} \gets 0\) for all \(c,k\).
\For{each class \(c\) in random order}
    \State \(g_k \gets \max\{q_k - n_k, 0\}\) for all \(k\)
    \State assign class \(c\) to the \(\rho_c\) clients with the largest \(g_k\)
    \For{each selected client \(k\)}
        \State \(M_{c,k} \gets 1\), \(n_k \gets n_k + 1\)
    \EndFor
\EndFor

\State \textbf{Image distribution:} for each class \(c\) with index set \(\mathcal{I}_c\), randomly permute \(\mathcal{I}_c\). For each client \(k\) such that \(M_{c,k}=1\), assign \(s\) non-overlapping images from \(\mathcal{I}_c\) to \(\mathcal{D}_k^\ell\).

\State \Return \(\mathbf{M}, \{\mathcal{D}_k^\ell\}_{k=1}^K\)
\end{algorithmic}
\end{algorithm}
\FloatBarrier
\subsection{Prevalence-based personalized weighted federated learning (PreP-WFL)}
\label{sec:prepwfl}

\begin{algorithm}[!ht]
\caption{PreP-WFL: Prevalence-based Personalized Weighted Federated Learning}
\label{alg:prep_wfl}
\begin{algorithmic}[1]
\Require Client set $\mathcal{K}=\{1,\dots,K\}$, class set $\mathcal{C}=\{1,\dots,C\}$;
labeled data $\{\mathcal{D}_k^{\ell}\}_{k=1}^K$ with
$\mathcal{D}_k^{\ell}=\{(x_i^{(k)},y_i^{(k)})\}_{i=1}^{n_k}$; \\
initial global encoder $\theta^{0}$, classifier head $\psi^{0}$; communication rounds $T$, local epochs $E$.
\Statex
\State \textbf{(Server) PreP-weight calculation}
\For{each client $k \in \mathcal{K}$}
    \State Client $k$ sends local label set $\mathcal{C}_k \subseteq \mathcal{C}$ (class IDs only)
\EndFor
\For{each class $c \in \mathcal{C}$}
    \State $\rho_c \gets \sum_{k=1}^K \mathbf{1}\{c \in \mathcal{C}_k\}$ \Comment{prevalence, Eq.~\eqref{eq:prevalence}}
    \State $w_c \gets 1/\rho_c$ \Comment{PreP-weight, Eq.~\eqref{eq:weight}}
\EndFor
\State Broadcast $\{w_c\}_{c=1}^C$ to all clients

\Statex
\State \textbf{(Server) Federated fine-tuning}
\State Initialize global parameters $\theta^{0}, \psi^{0}$
\For{$t = 1,\dots,T$}
    \State Broadcast $(\theta^{t}, \psi^{t})$ and $\{w_c\}$ to all clients
    \For{each client $k \in \mathcal{K}$ \textbf{in parallel}}
        \State $(\theta_k^{t+1}, \psi_k^{t+1}) \gets \textsc{ClientUpdate}(k,\theta^{t},\psi^{t},\{w_c\})$
    \EndFor
    \State $N_{\mathrm{tot}} \gets \sum_{k=1}^{K} n_k$
    \State $\theta^{t+1} \gets \sum_{k=1}^{K} \frac{n_k}{N_{\mathrm{tot}}}\,\theta_k^{t+1}$ 
    \State $\psi^{t+1}   \gets \sum_{k=1}^{K} \frac{n_k}{N_{\mathrm{tot}}}\,\psi_k^{t+1}$
\EndFor
\State \Return $(\theta^{T}, \psi^{T})$

\Statex
\Function{ClientUpdate}{$k,\theta^{t},\psi^{t},\{w_c\}$}
    \State $\theta_k \gets \theta^{t}$, $\psi_k \gets \psi^{t}$
    \State Build local weight vector $w^{(k)}$ by selecting $\{w_c : c \in \mathcal{C}_k\}$
    \For{each local epoch {$i = 1,\dots,E$} on $\mathcal{D}_k^{\ell}$}
        \For{each batch $\{(x_i^{(k)},y_i^{(k)})\}$ from $\mathcal{D}_k^{\ell}$}
            \State Compute $p_{\theta_k,\psi_k}(y\mid x)$
            \State Compute WCE loss $\mathcal{L}_k(\theta_k,\psi_k)$  \Comment{WCE loss, Eq.~\eqref{eq:wce}}
            \State Update $(\theta_k,\psi_k)$ by one step of Adam on $\mathcal{L}_k(\theta_k,\psi_k)$
        \EndFor
    \EndFor
    \State \Return $(\theta_k, \psi_k)$
\EndFunction

\end{algorithmic}
\end{algorithm}

In Fed-MAE fine-tuning, each client optimizes an unweighted
cross-entropy (CE) loss and thus treats all classes equally. In our diatom setting, however, some classes are site-specific and appear on only a few
clients. To explicitly account for such low-prevalence scenarios, we propose \textbf{PreP-WFL}, as illustrated in Fig~\ref{FIG:1}. The method adaptively reweights the local fine-tuning loss using cross-client class prevalence, statistically estimated from uploaded label IDs without sharing raw images or per-sample annotations. This amplifies rare, site-specific classes during optimization while keeping data local.

Specifically, let the client set be \(\mathcal{K}=\{1,\dots,K\}\) and the class set be \(\mathcal{C}=\{1,\dots,C\}\). Client \(k\) holds labeled data
\(\mathcal{D}_k^{\ell}\) with local label set \(\mathcal{C}_k\subseteq\mathcal{C}\).
In a one-shot statistics step, each client uploads only its local label set
\(\mathcal{C}_k\) (label IDs, no images) to the server. From these sets, the
server computes the prevalence of each class \(c\) as
\begin{equation}
\rho_c = \sum_{k=1}^K \mathbf{1}\{c \in \mathcal{C}_k\},
\label{eq:prevalence}
\end{equation}
i.e., the number of clients on which class \(c\) appears ($1 \le \rho_c \le K$). We then define a reciprocal prevalence-based class weight (PreP-weight):
\begin{equation}
w_c = \frac{1}{\rho_c}.
\label{eq:weight}
\end{equation}The server then broadcasts the weights \(\{w_c\}_{c=1}^C\) to all
clients, which construct local class-weight vectors by selecting the
entries corresponding to their own labels sets \(\mathcal{C}_k\).

On client \(k\), with labeled data
\(\mathcal{D}_k^{\ell}=\{(x_i^{(k)},y_i^{(k)})\}_{i=1}^{n_k}\), a local encoder
\(f_{\theta_k}\) and classifier head \(h_{\psi_k}\) define the predictive distribution
\(p_{\theta_k,\psi_k}(y \mid x) = \mathrm{softmax}\big(h_{\psi_k}(f_{\theta_k}(x))\big)_y\). We then replace CE loss with a prevalence-weighted loss:
\begin{equation}
\mathcal{L}_k(\theta_k,\psi_k)
= \frac{1}{n_k} \sum_{i=1}^{n_k}
\Big[-\,w_{y_i^{(k)}}\log p_{\theta_k,\psi_k}\big(y_i^{(k)} \mid x_i^{(k)}\big)\Big],
\label{eq:wce}
\end{equation}
which is implemented using PyTorch's built-in weighted cross-entropy loss (WCE) on each client. Federated aggregation remains based on FedAvg over \(\theta_k\) and \(\psi_k\). The overall procedure is summarized in Algorithm~\ref{alg:prep_wfl}.

\section{Experiments}
We adopt MAE and Fed-MAE as the SSL and SSFL framework, respectively. In all experiments, we use ViT-B~\cite{dosovitskiy2020image} as the backbone with a patch size of \(16 \times 16\).
\subsection{Pre-training setup}
We follow the MAE configuration with a masking ratio of \(75\%\) during pre-training. Optimization is performed with AdamW (\(\beta_1 = 0.9\), \(\beta_2 = 0.95\)). The base learning rate is set to \(1.5 \times 10^{-4}\). 
Federated pre-training is run for 800 communication rounds with a warm-up over the first 40 epochs and a batch size of 128. For data augmentation, we apply random resized cropping to generate input images of \(224 \times 224\) pixels, with the scale factor sampled from \([0.7, 1.0]\), followed by random horizontal and vertical flipping. 
\subsection{Fine-tuning setup}
During federated fine-tuning, the pre-trained encoder is further optimized for 100 communication rounds, with a warm-up period of 5 epochs and a batch size of 64. We employ layer-wise learning rate decay with a decay factor of 0.75 and use a base learning rate of \(1.0 \times 10^{-3}\). Input images are resized to \(224 \times 224\) pixels. We use ``\texttt{rand-m9-mstd0.5-inc1}'' augmentation policy~\cite{cubuk2020randaugment}, followed by random erasing (erasing probability 0.25, pixel-wise mode, one erased region per image).
\subsection{Evaluation metrics}
We use macro-averaged accuracy and F1-score as the evaluation metric to account for the strong class imbalance in the test set. Unlike micro-averaging, which is dominated by frequent classes, macro-averaging assigns equal weight to all classes, ensuring that rare diatom classes contribute equally to the final score. We evaluate all methods on the common test set $\mathcal{D}_{\text{test}}$ using the same 36-class global label space (closed-set evaluation).

\section{Results and Discussion}
This section presents a comprehensive evaluation of SSFL for diatom classification under diverse data distribution scenarios. We first establish baseline performance under ideal IID settings (\S\ref{sec:iid}), then systematically examine two key forms of non-IIDness: unlabeled-data-volume heterogeneity during pre-training (\S\ref{sec:pretrain_noniid}) and label-space heterogeneity during fine-tuning (\S\ref{sec:finetune_noniid}). Finally, we demonstrate the effectiveness of the proposed PreP-WFL in mitigating performance degradation in low-prevalence regimes (\S\ref{sec:prep_wfl}).

\subsection{SSFL performance under IID settings}
\label{sec:iid}
\begin{figure}
     \centering
     \includegraphics[width=.8\textwidth]{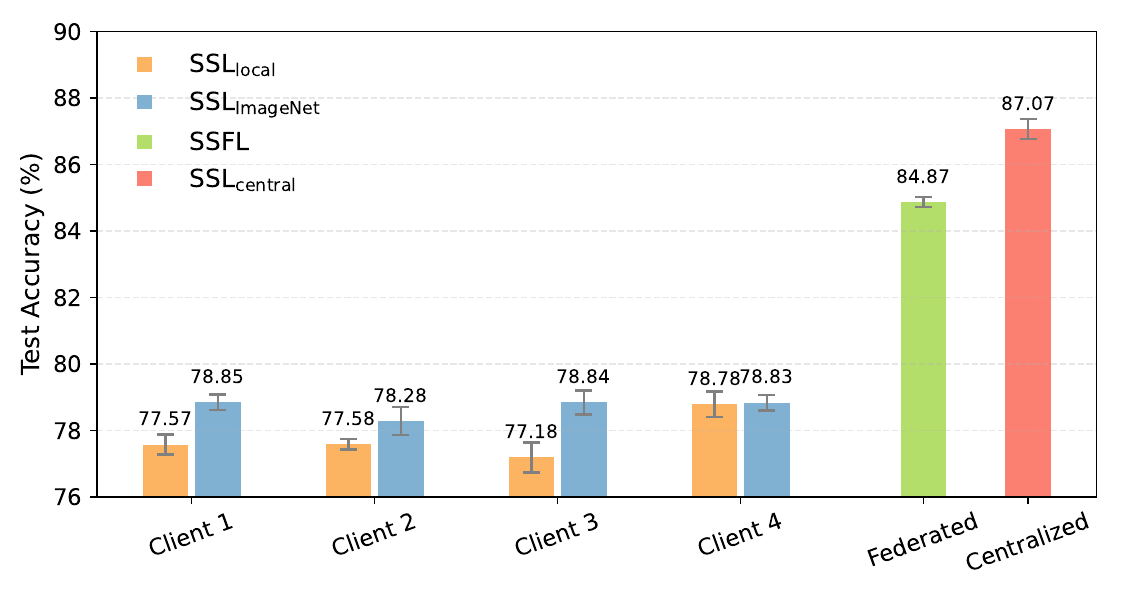}
     \caption{SSFL performance under IID settings (see also Table~\ref{tbl:iid_results}).}
     \label{fig:iid_results}
 \end{figure}

Under ideal IID settings, our results show that SSFL substantially outperforms isolated local training, while remaining close to centralized training performance, indicating that SSFL is an effective paradigm for diatom classification (Fig.~\ref{fig:iid_results}). 

The corresponding results are summarized in Table~\ref{tbl:iid_results}. In the local-only setting, each client $k$ either performs SSL pre-training on its own unlabeled split $\text{Split}^{u}_{k}$ or uses an encoder pre-trained on ImageNet~\cite{deng2009imagenet}, and then fine-tunes solely on its own labeled split $\text{Split}^{\ell}_{k}$. The resulting local accuracies lie in a range of approximately $77\text{--}79\%$. ImageNet initialization yields slightly higher accuracy than SSL pre-training on local data, likely due to the relatively small size of the local unlabeled datasets, consistent with the observations of Tan et al.~\cite{tan2025scaling}. In contrast, SSFL attains an accuracy of $84.87\%$ under the same IID distribution, clearly surpassing all individual local models. Centralized SSL, which pre-trains on the full unlabeled pool $\mathcal{D}^{u}$ and fine-tunes on the union of all labeled subsets $\bigcup_{k}\mathcal{D}_k^{\ell}$, achieves $87.07\%$. The corresponding F1-scores are reported in Table~\ref{apptab:iid} in Appendix~\ref{sec:appendix_results}.

These findings indicate that, when each client holds the same amount of unlabeled data and has access to all classes, the performance of isolated training is fundamentally constrained by limited sample diversity and scale. In contrast, SSFL enables institutions to share representation learning via model aggregation, effectively combining complementary data distributions and feature representations across sites. The small gap to the centralized model suggests that most of the benefits of data sharing can be realized through SSFL, without violating data-ownership concerns.

\begin{table}[width=.7\linewidth,cols=5,pos=h]
\caption{Accuracy results under IID settings. $\mathcal{D}_k^{u}$ and $\mathcal{D}_k^{\ell}$ denote the unlabeled and labeled datasets of client $k$, respectively.}
\label{tbl:iid_results}
\renewcommand{\arraystretch}{1.2}
\begin{tabularx}{\tblwidth}{@{} l c c c c @{}} 
\toprule
      & \textbf{Method} & \textbf{Pre-train data} & \textbf{Fine-tune data }& \textbf{Test Accuracy (\%)} \\
\midrule
\multirow{2}{*}{Client~1} 
  & \multirow{2}{*}{SSL} 
  & $\mathcal{D}^{u}_1$ in $\mathrm{Split}^{u}_{\mathrm{IID}}$
  & \multirow{2}{*}{$\mathcal{D}^{\ell}_1$ in $\mathrm{Split}^{\ell}_{\mathrm{IID}}$} 
  & 77.57 \\
  & 
  & ImageNet              
  & 
  & 78.85 \\
\midrule
\multirow{2}{*}{Client~2} 
  & \multirow{2}{*}{SSL} 
  & $\mathcal{D}^{u}_2$ in $\mathrm{Split}^{u}_{\mathrm{IID}}$
  & \multirow{2}{*}{$\mathcal{D}^{\ell}_2$ in $\mathrm{Split}^{\ell}_{\mathrm{IID}}$} 
  & 77.58 \\
  & 
  & ImageNet              
  & 
  & 78.28 \\
\midrule
\multirow{2}{*}{Client~3} 
  & \multirow{2}{*}{SSL} 
  & $\mathcal{D}^{u}_3$ in $\mathrm{Split}^{u}_{\mathrm{IID}}$
  & \multirow{2}{*}{$\mathcal{D}^{\ell}_3$ in $\mathrm{Split}^{\ell}_{\mathrm{IID}}$} 
  & 77.18 \\
  & 
  & ImageNet              
  & 
  & 78.84 \\
\midrule
\multirow{2}{*}{Client~4} 
  & \multirow{2}{*}{SSL} 
  & $\mathcal{D}^{u}_4$ in $\mathrm{Split}^{u}_{\mathrm{IID}}$
  & \multirow{2}{*}{$\mathcal{D}^{\ell}_4$ in $\mathrm{Split}^{\ell}_{\mathrm{IID}}$} 
  & 78.78 \\
  & 
  & ImageNet              
  & 
  & 78.83 \\
\midrule
Federated   & SSFL & Split$^{u}_{\text{IID}}$ & Split$^{\ell}_{\text{IID}}$ & \textbf{84.87} \\
\midrule
Centralized & SSL     & $\mathcal{D}^{u}$        & $\mathcal{D}^{\ell}$        & 87.07 \\
\bottomrule
\end{tabularx}
\end{table}

\subsection{SSFL performance under non-IID unlabeled data distributions}
\label{sec:pretrain_noniid}
\begin{figure}
    \centering
    \includegraphics[width=.6\textwidth]{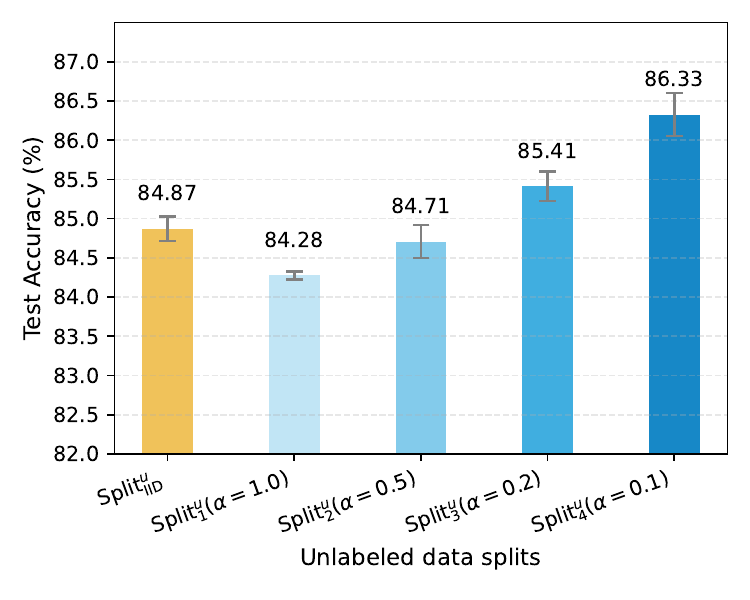}
    \caption{Results on unlabeled splits. Bars show mean accuracy and error bars denote standard deviation over three runs. More severe quantity skew in the unlabeled data (smaller $\alpha$) consistently yields higher accuracy.}
    \label{fig:pretrain_results}
\end{figure}

To isolate the effect of unlabeled-data-volume heterogeneity on representation learning, we maintain IID settings during fine-tuning ($\text{Split}^{\ell}_{\mathrm{IID}}$) while varying only the pre-training distribution across four heterogeneous unlabeled splits ($\text{Split}^{u}_{1}$--$\text{Split}^{u}_{4}$). Results displayed in Fig~\ref{fig:pretrain_results} show that, under FedAvg with data-size weighted aggregation, increasing quantity skew in the unlabeled data is associated with improved downstream performance after SSFL pre-training. The corresponding macro-F1 scores are reported in Table~\ref{apptab:pretrain_noniid} in Appendix~\ref{sec:appendix_results}.

Quantitatively, under mild to moderate heterogeneity ($\alpha=1.0$ or $0.5$), federated accuracies are slightly lower or comparable to the IID baseline (84.87\%), reaching 84.28\% and 84.71\%, respectively. With increasing heterogeneity, classification performance keeps improving: $\text{Split}^{u}_{3} (\alpha = 0.2)$ attains 85.41\% accuracy, and the most heterogeneous configuration, $\text{Split}^{u}_{4} (\alpha = 0.1)$ reaches 86.33\%, clearly surpassing the IID baseline. 

In centralized diatom SSL, enlarging the unlabeled dataset size is known to systematically improve diatom classification by capturing richer morphological structure and intra-class variability~\cite{tan2025scaling}. Our federated results extend this principle: under severe quantity skew, data-rich institutions contribute proportionally more gradient updates in FedAvg, steering the global encoder toward more generalizable feature representations. When heterogeneity is mild, the advantage of the largest client is insufficient to compensate for the limited diversity at smaller sites, yielding performance that is only comparable to, or slightly below, the IID case. As heterogeneity becomes more pronounced, the combination of richer morphology at data-abundant sites and volume-weighted aggregation increasingly dominates the optimization, leading to consistent accuracy gains.

\begin{table}[width=\linewidth,cols=2,pos=h]
\centering
\caption{Accuracy results on the most heterogeneous unlabeled split 
Split$^{u}_{4}$. For each client $k$, $\mathcal{D}^{u}_{k}$ and 
$\mathcal{D}^{\ell}_{k}$ denote its local unlabeled and labeled datasets, 
respectively.}
\label{tbl:split4}
\renewcommand{\arraystretch}{1.2}
\begin{tabular}{lccc}
\toprule
 & \textbf{Pre-train data} & \textbf{Fine-tune data} & \textbf{Test Accuracy (\%)} \\
\midrule
Client~1 & $\mathcal{D}^{u}_1$ in $\text{Split}^{u}_{4}$ & $\mathcal{D}_1^{\ell}$ in $\mathrm{Split}^{\ell}_{\mathrm{IID}}$
 & 79.69  \\
Client~2 & $\mathcal{D}^{u}_2$ in $\text{Split}^{u}_{4}$ & $\mathcal{D}_2^{\ell}$ in Split$^{\ell}_{\mathrm{IID}}
$ & 81.33  \\

Client~3 & $\mathcal{D}^{u}_3$ in $\text{Split}^{u}_{4}$ & $\mathcal{D}_3^{\ell}$ in Split$^{\ell}_{\text{IID}}$ & 70.64 \\
Client~4 & $\mathcal{D}^{u}_4$ in $\text{Split}^{u}_{4}$ & $\mathcal{D}_4^{\ell}$ in Split$^{\ell}_{\text{IID}}$ & 73.86 \\
\midrule
Federated & $\text{Split}^{u}_{4}$ &  $\mathrm{Split}^{\ell}_{\mathrm{IID}}$
 & 86.33  \\
\bottomrule
\end{tabular}
\end{table}

Table~\ref{tbl:split4} compares isolated local training with federated performance under the most heterogeneous unlabeled configuration, $\text{Split}^{u}_{4}$, and shows that the federated model (86.33\%) clearly outperforms all local models trained on the client side. The most data-scarce client (Client~3) exhibits a substantial improvement of roughly 16 percentage points, whereas the most data-abundant client (Client~2) still achieves a noteworthy gain of 5 points. This pattern indicates that both data-poor and data-rich institutions benefit from federated collaboration, as aggregation supplies complementary information that is absent from any single local dataset.

Overall, these findings indicate that unlabeled-data-volume heterogeneity need not be detrimental during SSFL federated pre-training when using FedAvg. Even under severe heterogeneity, collaboration with data-poor clients does not degrade the performance of data-rich institutions, but instead, their accuracy is further improved through federated aggregation. In practice, institutions with limited labeled data can still meaningfully contribute to and benefit from the federation by sharing large pools of unlabeled imagery, thereby strengthening the global representation.

\subsection{SSFL performance under non-IID labeled data distributions}
\label{sec:finetune_noniid}
\begin{figure}
    \centering
    \includegraphics[width=.75\textwidth]{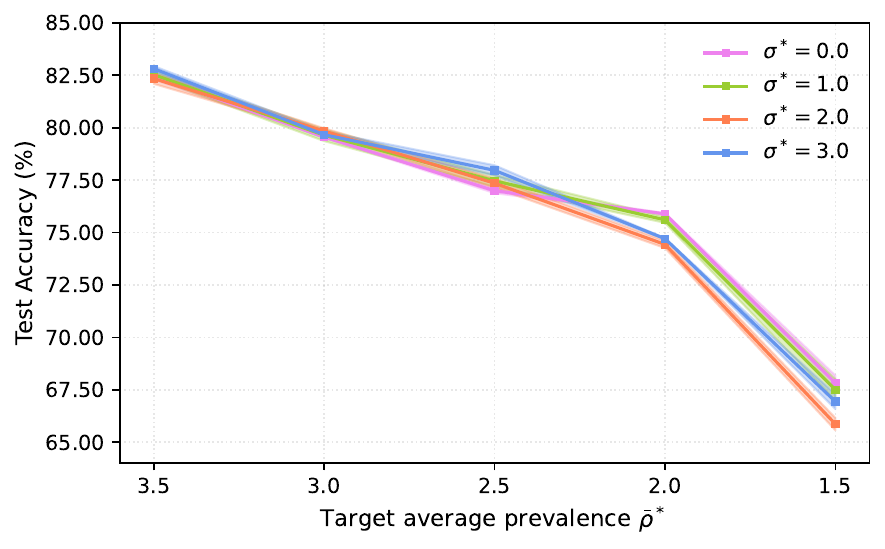}
    \caption{Results on non-IID labeled data.The x-axis shows the target average class prevalence $\bar{\rho}^*$ and the colored curves correspond to different levels of class-set size disparity $\sigma^*$. Accuracy is primarily driven by $\bar{\rho}$, whereas varying $\sigma$ leads to only minor differences across settings. See also Table~\ref{tab:finetune_noniid}.}
    \label{fig:finetune_noniid}
\end{figure}

To systematically characterize label-space heterogeneity during fine-tuning, we keep IID settings during pre-training ($\text{Split}^{u}_{\mathrm{IID}}$) while evaluating 20 non-IID labeled-data splits generated by our PreDi partitioning scheme. Figure~\ref{fig:finetune_noniid} reveals that class prevalence $(\rho)$ strongly determines SSFL fine-tuning performance.

Table~\ref{tab:finetune_noniid} reports the accuracy for all $(\bar \rho^*, \sigma^*)$ configurations. When $\sigma^*$ is fixed, federated accuracy decreases monotonically as $\bar\rho^*$ is reduced from $3.5$ to $1.5$, with a largest observed drop of $-16.48$ percentage points. In contrast, for fixed $\bar{\rho}^*$, varying $\sigma^*$ from
$0.0$ to $3.0$ induces only minor fluctuations. The corresponding F1-score results exhibit a similar trend and are reported in Table~\ref{apptab:finetune_noniid} in Appendix~\ref{sec:appendix_results}.

\FloatBarrier
\begin{table}[width=0.7\linewidth,cols=6,pos=h]
\caption{Accuracy results on non-IID labeled data under different ($\bar{\rho}^*,\sigma^*$) configurations.}
\label{tab:finetune_noniid}
\renewcommand{\arraystretch}{1.2}
\begin{tabularx}{\tblwidth}{@{} l *{5}{>{\centering\arraybackslash}X} @{}} 
\toprule
        & \multicolumn{5}{c}{\textbf{Test Accuracy (\%)}} \\ 
\cmidrule(lr){2-6}        
        & \(\bar\rho^*=3.5\) & \(\bar \rho^*=3.0\) & \(\bar \rho^*=2.5\)  & \(\bar \rho^*=2.0\)  & \(\bar \rho^*=1.5\) \\ 
\midrule
\(\sigma^*=0.0\) & 82.47 & 79.56 & 77.00 & 75.89 & 67.82 \\
\(\sigma^*=1.0\) & 82.53 & 79.64 & 77.47 & 75.61 & 67.50 \\
\(\sigma^*=2.0\) & 82.34 & 79.84 & 77.35 & 74.43 & 65.86 \\
\(\sigma^*=3.0\) & 82.81 & 79.67 & 77.97 & 74.71 & 66.94 \\ 
\bottomrule
\end{tabularx}
\end{table}

Table~\ref{tab:prevalence_disparity_a} compares the isolated local model with the federated model across prevalence levels at $\sigma^* = 0.0$. The federated model consistently outperforms all individual clients. Although its absolute accuracy decreases from $82.47\%$ to $67.82\%$ as $\bar{\rho}$ drops, but the margin over the local models grows substantially, from up to $+13.84$ percentage points at $\bar{\rho}^* = 3.5$ to as much as $+56.78$ points at $\bar{\rho}^* = 1.5$. This indicates that federated collaboration becomes increasingly beneficial in regimes with more low-prevalence, institution-specific classes. Table~\ref{tab:prevalence_disparity_b} compares isolated local models with the federated model across disparity levels at $\bar{\rho}^* =3.5$, Federated accuracy remains essentially stable (82.34–82.81\%) while consistently surpassing all individual clients. The widening gap to the weakest client primarily reflects local degradation rather than changes in the global federated model. Overall, these results reinforce that SSFL performance is primarily governed by cross-client class prevalence, whereas class-set size disparity can be largely compensated by federated aggregation.

\begin{table}[t]
\centering
\renewcommand{\arraystretch}{1.2}
\caption{Fine-tuning accuracy results under fixed disparity $\sigma^* = 0.0$ for varying prevalence $\bar{\rho}^*$.}
\label{tab:prevalence_disparity_a}
\begin{tabular}{lccccc}
\toprule
\multirow{2}{*}{\(\sigma^*=0.0\)} & \multicolumn{5}{c}{\textbf{Test Accuracy (\%)}} \\
\cmidrule(lr){2-6}
& \(\bar \rho^*=3.5\) & \(\bar \rho^*=3.0\) & \(\bar \rho^*=2.5\) & \(\bar \rho^*=2.0\) & \(\bar \rho^*=1.5\) \\
\midrule
Client~1 & 69.55 & 59.79 & 50.18 & 42.08 & 11.04 \\
Client~2 & 71.12 & 58.74 & 54.46 & 43.11 & 33.71 \\
Client~3 & 68.63 & 67.66 & 50.28 & 38.30 & 30.50 \\
Client~4 & 69.04 & 62.43 & 50.54 & 42.41 & 30.10 \\
\midrule
Federated & 82.47 & 79.56 & 77.00 & 75.89 & 67.82 \\
\bottomrule
\end{tabular}
\end{table}

\begin{table}[t]
\centering
\renewcommand{\arraystretch}{1.2}
\caption{Fine-tuning accuracy results under fixed prevalence $\bar{\rho}^* = 3.5$ for varying disparity $\sigma^*$.}
\label{tab:prevalence_disparity_b}
\begin{tabular}{lcccc}
\toprule
\multirow{2}{*}{\(\bar\rho^*=3.5\)} & \multicolumn{4}{c}{\textbf{Test Accuracy (\%)}} \\
\cmidrule(lr){2-5}
& \(\sigma^*=0.0\) & \(\sigma^*=1.0\) & \(\sigma^*=2.0\) & \(\sigma^*=3.0\) \\
\midrule
Client~1 & 69.55 & 69.37 & 71.96 & 72.24 \\
Client~2 & 71.12 & 65.85 & 64.40 & 62.38 \\
Client~3 & 68.63 & 68.71 & 66.34 & 66.48 \\
Client~4 & 69.04 & 74.16 & 75.60 & 77.90 \\
\midrule
Federated & 82.47 & 82.53 & 82.34 & 82.81 \\
\bottomrule
\end{tabular}
\end{table}

In particular, under low-prevalence settings, many classes become site-specific or nearly site-specific, so only a few clients contribute gradient updates for a given class during federated fine-tuning. This reduces both the effective sample diversity and the number of independent optimization trajectories per class, making the global model more vulnerable to local biases and overfitting. In contrast, when prevalence is high, each class is observed and updated by many clients, allowing the global model to aggregate diverse evidence and more stable gradients. By comparison, redistributing how many classes each client covers has only a limited effect as long as the prevalence pattern remains fixed, since it mainly permutes which clients are class-rich or class-poor, without substantially changing how many clients contribute to each individual class.

\subsection{Effectiveness of PreP-WFL in low-prevalence regimes}
\label{sec:prep_wfl}
\begin{figure}
    \centering
    \includegraphics[width=0.6\textwidth]{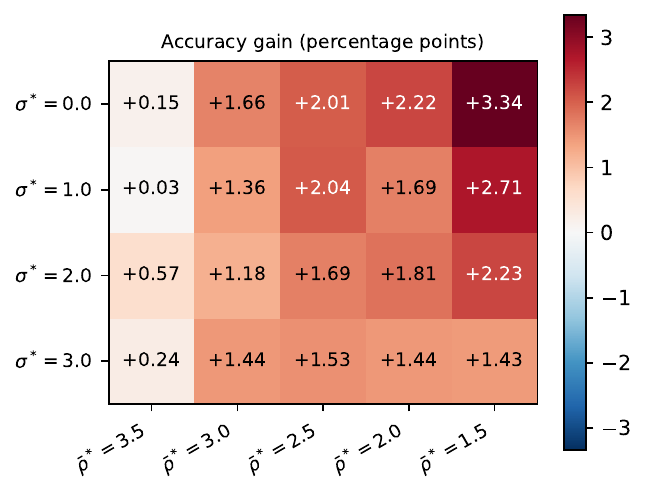}
    \caption{Accuracy gain (percentage points) of PreP-WFL over the Fed-MAE baseline across $(\bar{\rho}^*,\sigma^*) $ configurations of labeled data. }
    \label{fig:prep_gain}
\end{figure}

Given the dominant influence of class prevalence on federated performance (\S\ref{sec:finetune_noniid}), low-prevalence regimes expose a key limitation that only a small subset of clients provides supervised updates for a site-specific class. As a result, class-specific gradients become sparse and easily dominated by updates from widely shared classes, causing systematic underfitting of rare, institution-specific taxa. To address this issue, we propose \textbf{PreP-WFL} (\textbf{Pre}valence-based \textbf{P}ersonalized \textbf{W}eighted \textbf{F}ederated \textbf{L}earning) to specifically mitigate performance degradation in low-prevalence regimes. In particular, PreP-WFL personalizes local fine-tuning by reweighting each class-specific loss according to its global cross-client prevalence, statistically estimated from uploaded label IDs without sharing raw images or per-sample annotations. Consequently, site-specific classes with low prevalence receive larger weights, amplifying their learning signal during local updates and rebalancing the global optimization to avoid under-learning of rare classes. Figure~\ref{fig:prep_gain} presents the accuracy gains of PreP-WFL over the SSFL baseline (Fed-MAE) across all 20 PreDi configurations, showing a clear trend that performance improvements increase as class prevalence decreases.

\begin{table}[width=.8\linewidth,cols=7]
\caption{Accuracy results of Fed-MAE (baseline) and Fed-MAE with PreP-WFL (ours) across $(\bar{\rho}^*,\sigma^*)$ configurations.}
\label{tab:prep-wfl}
\renewcommand{\arraystretch}{1.2}
\begin{tabularx}{\tblwidth}{@{} ll *{5}{>{\centering\arraybackslash}X} @{}} 
\toprule
    & & \multicolumn{5}{c}{\textbf{Test Accuracy (\%)}} \\ 
\cmidrule(lr){3-7}         
    &  & \(\bar \rho^*=3.5\) & \(\bar \rho^*=3.0\) & \(\bar \rho^*=2.5\) & \(\bar \rho^*=2.0\) & \(\bar \rho^*=1.5\) \\ 
\midrule
\multirow{2}{*}{\(\sigma^*=0.0\)} 
    & Fed-MAE    
    & 82.47 & 79.56 & 77.00 & 75.89 & 67.82 \\
\cmidrule(lr){2-7}
& Fed-MAE$+$PreP-WFL (ours)
    & \textbf{82.62} & \textbf{81.22} & \textbf{79.01} & \textbf{78.11} & \textbf{71.16} \\
\midrule
\multirow{2}{*}{\(\sigma^*=1.0\)} 
    & Fed-MAE       
    & 82.53 & 79.64 & 77.47 & 75.61 & 67.50 \\ 
\cmidrule(lr){2-7}
    & Fed-MAE$+$PreP-WFL (ours)
    & \textbf{82.56} & \textbf{81.00} & \textbf{79.51} & \textbf{77.30} & \textbf{70.21} \\
\midrule
\multirow{2}{*}{\(\sigma^*=2.0\)} 
    & Fed-MAE      
    & 82.34 & 79.84 & 77.35 & 74.43 & 65.86 \\ 
\cmidrule(lr){2-7}
    & Fed-MAE$+$PreP-WFL (ours)
    & \textbf{82.91} & \textbf{81.02} & \textbf{79.04} & \textbf{76.23} & \textbf{68.09} \\
\midrule
\multirow{2}{*}{\(\sigma^*=3.0\)} 
    & Fed-MAE   
    & 82.81 & 79.67 & 77.97 & 74.71 & 66.94 \\ 
\cmidrule(lr){2-7}
    & Fed-MAE$+$PreP-WFL (ours)
    & \textbf{83.05} & \textbf{81.11} & \textbf{79.50} & \textbf{75.95} & \textbf{68.37} \\
\bottomrule
\end{tabularx}
\end{table}

Table~\ref{tab:prep-wfl} reports the performance of the Fed-MAE baseline and Fed-MAE with the proposed PreP-WFL across all $(\bar{\rho}^*,\sigma^*)$ configurations. At high prevalence ($\bar{\rho}^*= 3.5$), where most classes are observed by 3–4 clients, PreP-WFL yields only modest gains of $+0.03$ to $+0.57$ percentage points. This is expected: when prevalence is high, each class already receives abundant and diverse gradient updates under aggregation, so additional reweighting brings limited benefit. As prevalence decreases, the advantages of PreP-WFL become progressively more pronounced. By upweighting rare classes, PreP-WFL emphasizes their contribution during local fine-tuning and partially rebalances the global optimization landscape. The largest improvement occurs in the most challenging prevalence setting ($\bar{\rho}^* = 1.5$), where accuracy increases from 67.82\% (baseline) to 71.16\% with PreP-WFL. This corresponds to an approximate 22\% reduction in the performance gap between high- and low-prevalence settings (from 
82.47\% vs.\ 67.82\% to 82.62\% vs.\ 71.16\%), demonstrating that PreP-WFL substantially improves the low-prevalence scenarios. 

Moreover, within each prevalence level, PreP-WFL's performance gains systematically diminish as disparity $\sigma$ increases. For instance, at $\bar{\rho}^* = 1.5$, the improvement shrinks from $+3.34$ points at $\sigma^* =0.0$ to $+1.43$ points at $\sigma^* =3.0$. This trend is consistent across low-prevalence settings, as extreme disparity concentrates many classes in a few clients, so that a globally rare class can be locally common at a particular site. In such cases, aggressive up-weighting based on global prevalence becomes suboptimal, because the class is not locally rare and may already receive sufficient optimization emphasis. The corresponding F1-scores exhibit a similar trend and are reported in Table~\ref{apptab:prep_wfl} in Appendix~\ref{sec:appendix_results}.

In summary, these results indicate that PreP-WFL provides an effective mechanism to recover a substantial portion of the performance lost in low-prevalence federated regimes. By relying only on aggregated label statistics, the approach remains data-local and can be seamlessly integrated into existing SSFL pipelines. This highlights its practical value for real-world deployments in which rare, site-specific classes are of primary importance.

\section{Conclusion}
This work studies label-scarce visual classification under decentralized and 
heterogeneous data through the lens of SSFL. Using diatom classification as a 
representative real-world setting, we demonstrate that SSFL constitutes an effective data-local learning paradigm that consistently outperforms local-only training under both homogeneous and heterogeneous conditions. By explicitly decoupling heterogeneity across pre-training and fine-tuning, we further show that conflating these stage-specific data heterogeneities obscures their individual effects. For label-space heterogeneity specifically, we introduce PreDi, a partitioning scheme that disentangles prevalence and disparity into orthogonal dimensions, enabling controlled analysis of their individual effects. Experiments reveal that prevalence is the primary factor governing downstream performance, whereas disparity plays a comparatively minor role. Building on this finding, we propose PreP-WFL, a prevalence-based personalized weighting strategy that improves robustness in low-prevalence regimes, with gains increasing as prevalence decreases. 

Overall, these contributions provide a mechanistic basis for characterizing stage-specific data heterogeneity in SSFL and offer principled insights for designing data-local recognition systems in heterogeneous decentralized environments. One limitation of this study is that the analysis is conducted on a representative decentralized setting with a fixed federation scale.  Future work will extend this framework to larger and more diverse federations, jointly varying multiple sources of heterogeneity.

\section{Declaration of Competing Interest}
The authors declare no competing interests.
\section{CRediT Author Statement}
\textbf{Mingkun Tan:} Conceptualization, Methodology, Software, Validation, Formal analysis, Investigation, Data Curation, Writing - Original Draft, Visualization, Project administration. \textbf{Xilu Wang:} Conceptualization, Formal analysis, Writing - Review and Editing, Supervision. \textbf{Michael Kloster:} Resources, Data Curation, Writing - Review and Editing. \textbf{Tim W. Nattkemper:} Conceptualization, Resources, Writing - Review and Editing, Supervision, Project administration, Funding acquisition.

\section{Acknowledgement}
This work was supported by the de.NBI Cloud within the German Network for Bioinformatics Infrastructure (de.NBI) and ELIXIR-DE (Forschungszentrum Jülich and W-de.NBI-001, W-de.NBI-004, W-de.NBI-008, W-de.NBI-010, W-de.NBI-013, W-de.NBI-014, W-de.NBI-016, W-de.NBI-022).
\subsection{Funding}
M.T. and M.K. were funded by the Deutsche Forschungsgemeinschaft (DFG, German Research Foundation; project number: 463395318, GZ: NA 731/11-1).

\section{Declaration of generative AI and AI-assisted technologies in the manuscript preparation process}
During the preparation of this work the authors used ChatGPT in order to improve the readability and language of the manuscript. After using it, the authors reviewed and edited the content as needed and took full responsibility for the content of the published article.
\appendix
\section{Appendix}
\label{sec:appendix_results}

\begin{table}[width=.7\linewidth,cols=5,pos=h]
\caption{F1-score results under IID settings. $\mathcal{D}_k^{u}$ and $\mathcal{D}_k^{\ell}$ denote the unlabeled and labeled datasets of client $k$, respectively.}
\label{apptab:iid}
\renewcommand{\arraystretch}{1.2}
\begin{tabularx}{\tblwidth}{@{} l c c c c @{}} 
\toprule
      & \textbf{Method} & \textbf{Pre-train data} & \textbf{Fine-tune data }& \textbf{Test F1-score} \\
\midrule
\multirow{2}{*}{Client~1} 
  & \multirow{2}{*}{SSL} 
  & $\mathcal{D}^{u}_1$ in $\mathrm{Split}^{u}_{\mathrm{IID}}$
  & \multirow{2}{*}{$\mathcal{D}^{\ell}_1$ in $\mathrm{Split}^{\ell}_{\mathrm{IID}}$} 
  & 0.7054 \\
  & 
  & ImageNet              
  & 
  & 0.7234 \\
\midrule
\multirow{2}{*}{Client~2} 
  & \multirow{2}{*}{SSL} 
  & $\mathcal{D}^{u}_2$ in $\mathrm{Split}^{u}_{\mathrm{IID}}$
  & \multirow{2}{*}{$\mathcal{D}^{\ell}_2$ in $\mathrm{Split}^{\ell}_{\mathrm{IID}}$} 
  & 0.7100 \\
  & 
  & ImageNet              
  & 
  & 0.7186 \\
\midrule
\multirow{2}{*}{Client~3} 
  & \multirow{2}{*}{SSL} 
  & $\mathcal{D}^{u}_3$ in $\mathrm{Split}^{u}_{\mathrm{IID}}$
  & \multirow{2}{*}{$\mathcal{D}^{\ell}_3$ in $\mathrm{Split}^{\ell}_{\mathrm{IID}}$} 
  & 0.7051 \\
  & 
  & ImageNet              
  & 
  & 0.7552 \\
\midrule
\multirow{2}{*}{Client~4} 
  & \multirow{2}{*}{SSL} 
  & $\mathcal{D}^{u}_4$ in $\mathrm{Split}^{u}_{\mathrm{IID}}$
  & \multirow{2}{*}{$\mathcal{D}^{\ell}_4$ in $\mathrm{Split}^{\ell}_{\mathrm{IID}}$} 
  & 0.7203 \\
  & 
  & ImageNet              
  & 
  & 0.7215 \\
\midrule
Federated   & SSFL & Split$^{u}_{\text{IID}}$ & Split$^{\ell}_{\text{IID}}$ & 0.8008 \\
\midrule
Centralized & SSL     & $\mathcal{D}^{u}$        & $\mathcal{D}^{\ell}$        & 0.8236 \\
\bottomrule
\end{tabularx}
\end{table}

 \begin{table}[width=.5\linewidth,cols=2,pos=h]
 \caption{F1-score results on unlabeled splits.}
 \label{apptab:pretrain_noniid}
 \renewcommand{\arraystretch}{1.2}
 \begin{tabularx}{\tblwidth}{@{} l >{\centering\arraybackslash}X @{}} 
\toprule
\textbf{Pre-train data} &  \textbf{Test F1-score}  \\
\midrule
$\text{Split}^{u}_{\text{IID}}$ & 0.8008 \\
\midrule
$\text{Split}^{u}_{1} (\alpha =1.0)$ & 0.7934 \\
$\text{Split}^{u}_{2} (\alpha =0.5)$   & 0.7975 \\
$\text{Split}^{u}_{3} (\alpha =0.2)$    & 0.8063 \\
$\text{Split}^{u}_{4} (\alpha =0.1)$    & 0.8168 \\
\bottomrule
\end{tabularx}
\end{table}

\begin{table}[width=0.8\linewidth,cols=6,pos=h]
\caption{F1-score results on non-IID labeled data under different ($\bar{\rho}^*,\sigma^*$) configurations.}
\label{apptab:finetune_noniid}
\renewcommand{\arraystretch}{1.2}
\begin{tabularx}{\tblwidth}{@{} l *{5}{>{\centering\arraybackslash}X} @{}} 
\toprule
        & \multicolumn{5}{c}{\textbf{Test F1-score}} \\ 
\cmidrule(lr){2-6}        
        & \(\bar \rho^*=3.5\) & \(\bar \rho^*=3.0\) & \(\bar \rho^*=2.5\)  & \(\bar \rho^*=2.0\)  & \(\bar \rho^*=1.5\) \\ 
\midrule
\(\sigma^*=0.0\) & 0.7853 & 0.7570 & 0.7310 & 0.7217 & 0.6277 \\
\(\sigma^*=1.0\) & 0.7833 & 0.7589 & 0.7357 & 0.7163 & 0.6244 \\
\(\sigma^*=2.0\) & 0.7827 & 0.7611 & 0.7312 & 0.7059 & 0.6010 \\
\(\sigma^*=3.0\) & 0.7882 & 0.7579 & 0.7404 & 0.7038 & 0.6158 \\  
\bottomrule
\end{tabularx}
\end{table}
\begin{table}[width=.8\linewidth,cols=7,pos=h]
\caption{F1-score results of Fed-MAE (baseline) and Fed-MAE with PreP-WFL (ours) across $(\bar{\rho}^*,\sigma^*)$ configurations.}
\label{apptab:prep_wfl}
\renewcommand{\arraystretch}{1.2}
\begin{tabularx}{\tblwidth}{@{} ll *{5}{>{\centering\arraybackslash}X} @{}} 
\toprule
    & & \multicolumn{5}{c}{\textbf{Test F1-score}} \\ 
\cmidrule(lr){3-7}         
    &  & \(\bar\rho^*=3.5\) & \(\bar \rho^*=3.0\) & \(\bar\rho^*=2.5\) & \(\bar\rho^*=2.0\) & \(\bar\rho^*=1.5\) \\ 
\midrule
\multirow{2}{*}{\(\sigma^*=0.0\)} 
    & Fed-MAE    
    & 0.7853 & 0.7570 & 0.7310 & 0.7217 & 0.6277 \\
\cmidrule(lr){2-7}
    & Fed-MAE$+$PreP-WFL (ours)
    & \textbf{0.7822} & \textbf{0.7696} & \textbf{0.7426} & \textbf{0.7364} & \textbf{0.6611} \\
\midrule
\multirow{2}{*}{\(\sigma^*=1.0\)} 
    & Fed-MAE       
     & 0.7833 & 0.7589 & 0.7357 & 0.7163 & 0.6244\\ 
\cmidrule(lr){2-7}
    & Fed-MAE$+$PreP-WFL (ours)
    & \textbf{0.7838} & \textbf{0.7687} & \textbf{0.7492} & \textbf{0.7279} & \textbf{0.6473} \\
\midrule
\multirow{2}{*}{\(\sigma^*=2.0\)} 
    & Fed-MAE      
    & 0.7827 & 0.7611 & 0.7312 & 0.7059 & 0.6010 \\ 
\cmidrule(lr){2-7}
    & Fed-MAE$+$PreP-WFL (ours)
    & \textbf{0.7842} & \textbf{0.7689} & \textbf{0.7420} & \textbf{0.7179} & \textbf{0.6115} \\
\midrule
\multirow{2}{*}{\(\sigma^*=3.0\)} 
    & Fed-MAE   
    & 0.7882 & 0.7579 & 0.7404 & 0.7038 & 0.6158 \\ 
\cmidrule(lr){2-7}
    & Fed-MAE$+$PreP-WFL (ours)
    & \textbf{0.7880} & \textbf{0.7674} & \textbf{ 0.7472} & \textbf{0.7088} & \textbf{0.6173} \\
\bottomrule
\end{tabularx}
\end{table}

\FloatBarrier
\printcredits

\bibliographystyle{model1-num-names}

\bibliography{cas-refs}


\end{document}